# Fine-Tuned Small Language Model for Synthetic Clinical Data Generation and Predictive Healthcare Analytics

***Mihnea-Sebastian Burlacu, Ionuț Anghel*, Tudor Cioara***
*Computer Science Department*
*Technical University of Cluj-Napoca*
*Cluj-Napoca, Romania*
*burlacu.pa.mihnea@student.utcluj.ro, ionut.anghel@cs.utcluj.ro, tudor.cioara@cs.utcluj.ro*

**Abstract:** The reduced availability of medical datasets represents an important challenge in the development of predictive models for remote healthcare solutions, especially when the available data are imbalanced or difficult to extend through traditional methods. Small Language Models (SLMs) have become a promising alternative to Large Language Models (LLMs) for addressing these limitations due to their lower computational requirements and adaptability to specialized tasks. However, their performance remains highly dependent on the quality and representativeness of the available training data. In this paper, we propose a method for clinical data augmentation based on a fine-tuned Microsoft Phi-2 model that is used to generate synthetic instances, with the purpose of improving deep learning-based model for predicting remotely monitored vital signs. The Phi-2 model was adapted to the medical domain through supervised fine-tuning on the MedQuad dataset, through a Quantized Low-Rank Adaptation (QLoRA)-based method and by applying few-shot prompting techniques. The resulting model was used afterwards in the augmentation of a small-sized dataset, composed of vital sign values and labels associated with pathological and physiological states. The experimental results demonstrate that fine-tuned SLMs improved the performance on medical domain, achieving higher scores for both lexical (ROUGE) and semantic (BERTScore) evaluation metrics compared to the pretrained model. Moreover, the fine-tuned SLM is effective for generating structured synthetic clinical data. When used for data augmentation, the generated dataset improved the performance of a LSTM predictive model, increasing the classification accuracy from 89.0% to 93.7% and showing consistent improvements in precision, recall, and F1-score.

# Introduction

Remote healthcare solutions rely on predictive models to support the monitoring and early detection of physiological and pathological conditions (Abedi et al., 2025). However, the development of accurate machine learning models in healthcare is affected by the limited availability of clinical datasets, which are frequently small, imbalanced, and difficult to extend due to privacy, ethical, and regulatory constraints (Salmi et al., 2024), (Mennella et al., 2024). Recent advances in Large Language Models (LLMs) have demonstrated remarkable capabilities in processing medical information, generating clinical text, and supporting healthcare-related tasks (Artsi et al., 2025). However, their successful usage in the medical domain requires not only advanced capabilities in natural language generation, but also domain-specific adaptation, specific enhancements, thorough evaluation and the consideration of aspects such as confidentiality, bias, clinical safety and robustness (Al-Garadi et al., 2025), because errors or hallucinations can influence clinical interpretation and the decision-making process.

Recent research has explored the adaptation of LLMs to the medical domain through fine-tuning processes on specialized datasets, becoming more suitable for a particular task or area of application (Anisuzzaman et al., 2025). Through exposure to question-answer examples, doctor-patient dialogues or clinical instructions, language models can acquire capabilities of understanding clinical terminology and generating text like medically specific communication, which enables them to provide relevant clinical information. However, the use of LLMs assumes considerable computational resources, which can make their health domain adaptation and execution on environments with limited resources, rather challenging (Kim et al., 2025). Small Language Models (SLMs) are a promising alternative, because they have lower computational requirements, thus being more easily adaptable to specialized tasks and more suitable for local processing or use on edge devices (Lu et al., 2025).

*Corresponding author

At the same time an important challenge within healthcare systems is represented by the reduced availability of quality training data, medical datasets being often difficult to access or use because of confidentiality requirements, costs or restricted access, this affecting the reproducibility and generalization of research in this domain (Alberto et al., 2023). In addition, some of the open access medical datasets are reduced in size, incomplete, imbalanced, noisy or heterogeneous which can limit the performance of models that use machine learning or deep learning techniques to process clinical information (Miotto et al., 2018). Thus, data augmentation techniques are frequently used to address this problem, the generation of synthetic data representing an important direction for extending and diversifying these training datasets. This constitutes a new and relevant use case for LLMs/SLMs due to their capacity to generate coherent content that allows the augmentation of existing data or even the generation of new datasets, with statistical properties close to those of real clinical data (Barr et al., 2025). However, even though the use of LLMs as a generative tool for the augmentation of a structured clinical dataset is tackled in the literature (Hsieh et al., 2025), (Montenegro et al., 2025) limited attention has been given to the use of domain-adapted Small Language Models (SLMs) for the same purpose. There is a lack of studies evaluating whether SLM-generated synthetic data can effectively improve the performance of downstream predictive models while maintaining the computational efficiency advantages that make SLMs attractive for healthcare applications.

To address these limitations, this paper investigates the use of a domain-adapted Small Language Model (SLM) as a generative component for structured clinical data augmentation and evaluates its impact on downstream predictive modeling tasks.

First, the pretrained Microsoft Phi-2 SLM (Javaheripi et al., 2023) was adapted to the medical domain using the Medical Question Answering Dataset (MedQuad) (Ben Abacha & Demner-Fushman, 2019) and a supervised fine-tuning process based on Quantized Low-Rank Adaptation (QLoRA). The resulting model was used to generate synthetic samples for augmenting a limited dataset composed of vital sign values and labels associated with physiological or pathological states. Finally, the augmented dataset is used to retrain and optimize a Long Short-Term Memory (LSTM) predictive model (Burlacu & Anghel, 2024) enabling the assessment of whether SLM-generated clinical data can improve prediction performance in remote healthcare scenarios.

The main contributions of this paper are the following:

- The design of a medical data augmentation pipeline based on a fine-tuned SLM
- The adaptation of the pretrained Phi-2 model on the MedQuad dataset, using QLoRA, and its use for generating synthetic data samples, through few-shot prompting.
- The evaluation of the impact of the synthetic data on the performance of an LSTM predictive model considering diverse augmentation scenarios.

The rest of the paper is structured as follows: Related Work section presents the state of the art for SLMs, fine-tuning of medical language models and data augmentation; Methods section describes the proposed method for the fine-tuning process, the data augmentation strategy and the retraining of the LSTM model; Results and discussion sections shows the results obtained for the two models and Conclusion section presents the conclusions of the paper and possible future research directions.

## Related work

The recent development of language models has determined an increasing interest for their use in various domains, including in the medical domain, due to their capacity to process and generate natural language. Although pretrained language models have demonstrated strong capabilities of understanding human language and solving complex tasks through text generation, this being possible due to the fact that they were trained on very large volumes of textual data (Zhao et al., 2023), they were not explicitly trained to capture particularities from the medical domain, such as specialized terminology, clinical reasoning patterns or the specific behaviour of medical personnel oriented towards patients. Multiple studies have explored the adaptation of both LLMs and SLMs to the medical domain through fine-tuning techniques to improve the capacity of the models to understand medical terminology, to generate more relevant responses so that they can be used in various tasks specific to healthcare, such as consultations, diagnoses, patient management, etc. (Anisuzzaman et al., 2025). LLMs can aid the analysis pipelines of the values of vital signs collected through wearable devices or Internet of Things (IoT) systems, using machine learning and deep learning methods. LSTM models can be used to estimate the risk of clinical deterioration of patients, based on the continuous monitoring of several vital signs, among which are heart rate, respiratory rate, temperature and oxygen saturation (Scheid et al., 2025). Similarly, transformer models can predict the values of several vital signs, such as blood pressure, heart rate, temperature, oxygen saturation and respiratory rate for patients in intensive care (He et al., 2025). However, such approaches rely on comprehensive processing and understanding of the datasets, thus LLMs can be used as a pre-training step to analyse features, augment data and create insights.

The literature highlights the fact that the *applications of language models in the medical domain* can be viewed from the perspective of the users, (Nazi & Peng, 2024) grouping these applications into three categories: applications oriented towards patients, such as virtual medical assistants and health chatbots, applications oriented towards physicians, such as clinical decision support and medical diagnosis and applications oriented towards medical materials and research, such as automatic synthesis of medical reports. Starting from these general categories of applications, works can also be analysed depending on the purpose for which the models are adapted, the method used for adaptation and the type of data used in this process. A first direction is represented by the development of medical conversational assistants with the purpose of adapting language models to medical personnel – patient interactions and to medical question answering tasks, for which the model must understand the questions formulated in natural language and generate medically relevant responses, this direction being highlighted in the literature as an important scenario in the use of language models for healthcare domain (Wang & Zhang, 2024). A relevant example for this direction is ChatDoctor (Li et al., 2023), a language model based on LLaMA, fine-tuned on real doctor-patient conversations for which the authors collected approximately 100 000 dialogues, and they used another set of 10 000 instances for evaluation, such that the obtained model is oriented towards conversational consultation. To reduce the risk of hallucinations, in addition to the fine-tuning process, ChatDoctor also integrates a Knowledge Brain component that retrieves relevant medical information from external sources, both online and offline, among them also being a medical database composed of information about various pathologies, symptoms, tests, medicines and treatments. In addition to conversational medical assistants, other works have focused on adapting models to clinical tasks to help healthcare professionals, demonstrating that fine-tuning processes can also help in medical decision support, not only in generating responses to medical questions. LlamaCare (Li et al., 2024) proposes a Llama 2 model adapted for the clinical NLP domain, for which the authors built an instruction-input-output dataset, in which experts from the medical domain formulated the instructions and the inputs and outputs were extracted from clinical notes taken from the MIMIC-III benchmark. The model was adapted through instruction fine-tuning using LoRA and the evaluation included both clinical text generation for discharge summaries or support in radiology, as well as classification tasks, such as the prediction of clinical outcomes or of the length of stay in hospital. In the same direction is also included Clinical Camel (Toma et al., 2023), a medical language model adapted from Llama 2 using QLoRA, through which the authors introduced a Dialogue-Based Knowledge Encoding (DBKE) method that transforms dense medical texts into synthetic doctor – patient conversations, used for fine-tuning process. For the evaluation of the model, several medical benchmarks were used, such as MedQA, PubMedQA, MedMCQA and USMLE, thus highlighting the fact that the model can also perform the generation of clinical notes, not only responses to medical questions.

SLMs have attracted attention recently, because they can answer some limitations of LLMs, such as high memory and computation requirements, high latency, high costs and dependence of cloud infrastructures, these aspects being important in the medical domain, where the transmission of sensitive data to external servers can raise confidentiality problems. In this context, they are considered suitable for scenarios with limited resources, for local processing and use on edge devices, offering a more efficient alternative and easier to adapt to specialized tasks (Wang et al., 2025). RAD-PHI2 (Ranjit et al., 2024) can be included in this category, as it adapts the Microsoft Phi-2 model through instruction fine-tuning to clinical tasks specific to radiology, such as generating answers to questions about symptoms, imaging, differential diagnosis and treatment. For this, authors built a dataset formed both from question-answer pairs from Radiopaedia articles, as well as a set of instructions for tasks on Chest X-Ray radiology reports taken from MIMIC-CXR benchmark. MedMobile (Vishwanath et al., 2024) proposes the adaptation of Microsoft Phi-3 Mini, for medical purposes to obtain a model capable of running on mobile devices. This model was evaluated on USMLE benchmark and obtained a score of 75.7%, thus exceeding the minimum passing threshold and approaching the performances of much larger models, demonstrating that SLMs can be efficiently adapted for complex medical tasks.

Another relevant direction for the proposed work is the use of *language models for data augmentation*, which aims to extend and diversify the training datasets, without requiring the collection of new information. This direction has become increasingly important, because LLMs can generate new textual examples, labels, reformulations of structured data, being able to contribute to reducing the dependence on manually annotated data, which are often difficult and costly to obtain (Ding et al., 2024). Such approaches are especially important for situations in which the datasets are limited, and the use of other methods is constrained by confidentiality requirements or costs. (Ehrett et al., 2024) highlight the fact that these models can be used in generating new synthetic data to contribute for increasing the size and diversity of the training set, which could lead to the improvement of the performance of medical applications. MedSyn (Kumichev et al., 2024) uses a fine-tuned LLaMA 7B model for generating synthetic clinical notes using medical information associated with ICD-10 codes, these codes being the identifiers used by medical institutions to uniformly record diagnoses and health problems. Following this process, the authors obtained over 41 000 new synthetic data for 219 ICD-10 codes, which were later used to improve the performance of a classification model based on these codes. In a similar direction, (Lu et al., 2021) use a GPT-2 model fine-tuned on clinical notes from MIMIC-III for generating new data with the purpose of improving the performance of a model that estimates the risk of patient hospital readmission.

Most of the above directions are predominantly investigated as independent solutions, with limited efforts toward their integration into a unified framework. Therefore, there is a gap in leveraging the complementary strengths of these approaches within a single methodology. To address this limitation, the proposed approach establishes a coherent workflow in which a language model is first fine-tuned on domain-specific medical knowledge. Moreover, the proposed method is superior to the traditional data augmentation, as it allows obtaining new values coherent both from a semantic point of view, as well as a clinical one. Classical methods such as noise injection, random over-sampling, synthetic minority over-sampling (SMOTE) or adaptive synthetic sampling (ADASYN) can alter the relationships between the physiological values and can produce a medically unrealistic combination between the physiological values because they do not always manage to capture the complex dependencies between all the vital signs. Our solution, however, can generate consistent data based on the existing clinical information and even to create a new medically relevant characteristics due to the contextual understanding of medical domain.

# Methods

Figure 1 illustrates the proposed SLM-based medical data augmentation pipeline, including model adaptation, synthetic data generation, and downstream predictive model evaluation.

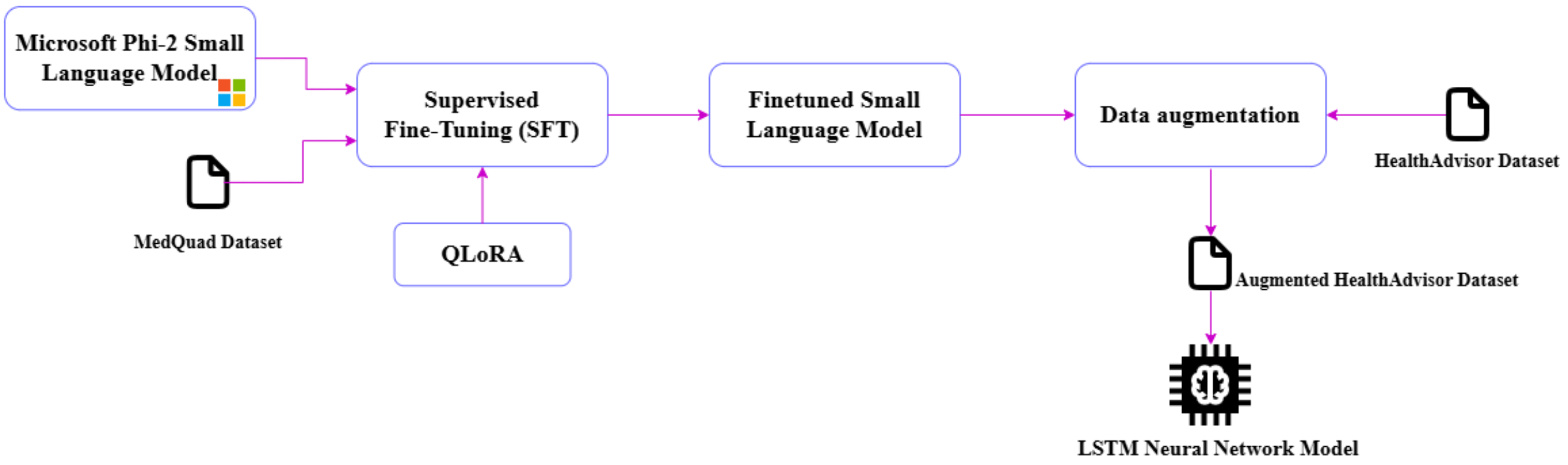

**Figure 1. SLM-based medical data augmentation pipeline**

A pre-trained Microsoft Phi-2 SLM model presented at NeurIPS 2023 (Javaheripi et al., 2023) was fine-tuned for the medical field using the MedQuad dataset (Ben Abacha & Demner-Fushman, 2019). Then, through few-shot prompt engineering, the model was used to augment the Health Advisor Dataset, by generating synthetic data samples. Finally the improved dataset was used to retrain a downstream LSTMprediction model to analyse the impact on model performance.

## *Fine-tuning the SLM*

Phi-2 is a compact 2.7-billion-parameter SLM designed to deliver strong reasoning and language understanding capabilities. Despite its relatively small size, it achieves competitive performance compared to significantly larger models, benefiting from advances in model scaling and high-quality training data curation. Its efficiency and accessibility make it well-suited for research applications and task-specific fine-tuning. To adapt the pre-trained Microsoft Phi-2 SLM model to the medical domain, such that it can generate values of the vital signs depending on the described health state for data augmentation, we performed a process of Supervised Fine-Tuning (SFT), using a Quantized Low-Rank Adaptation (QLoRA) (Dettmers et al., 2023) technique.

The fine-tuning of the model was performed using the dataset MedQuad that contains pairs of questions and answers from the medical domain. It is a curated collection of question–answer pairs compiled from 12 authoritative websites of the National Institutes of Health (NIH). The data spans a broad range of medical topics, including resources such as cancer.gov and the Genetic and Rare Diseases Information Center (GARD). Initially, these pairs were written in the form of a single text field that included both the instruction (the question) and the answer (see Fig. 2).

```
Instruction: What is (are) Parasites - Cysticercosis?

Response: Cysticercosis is an infection caused by the larvae of the parasite Taenia solium. This infection occurs after a person swallows tapeworm eggs. The larvae get into tissues such as muscle and brain, and form cysts there (these are called cysticerci). When cysts are found in the brain, the condition is called neurocysticercosis.
```

**Figure 2. MedQuad dataset sample**

The data was reformatted into instruction-answer pairs to support supervised fine-tuning. Each pair was subsequently structured into an explicit prompt template in which we have a short introductory context, the instruction section that contains the actual question, the output section that contains the corresponding answer to the question and a marker that indicates the end of the answer, allowing the model to learn the conversation structure that will be used later in the inference stage. After the formatting of the prompts, the data was tokenized using the tokenizer associated with the Phi-2 model, transforming the text into sequences of numerical tokens that can be processed by the model and the sequences longer than the maximum value accepted from the Phi-2 configuration were truncated. The resulting dataset was randomly shuffled to reduce possible biases and was split into two parts, 90% for training (training set) and 10% for evaluation (evaluation set).

The pre-trained Phi-2 SLM contains 2.7 billion parameters $W_0$ which represent the knowledge and language patterns acquired during the pretraining process:

$$W_0 \in R^{d \times k} \tag{1}$$

where $d$ represents the input dimension, $k$ represents the output dimension. These parameters enable the model to perform a wide range of natural language understanding and generation tasks and serve as the foundation for subsequent domain adaptation through fine-tuning.

Although Phi-2 is a rather compact SLM the fine-tuning process requires considerable hardware resources. To reduce the computational and memory requirements of the model while preserving the pretrained knowledge, we quantized the Phi-2 parameters from 16- or 32-bit floating-point values to 4-bit representations. The model weights were quantized using NormalFloat on 4 bits (NF4), while the internal forward pass computations are carried out in float16, maintaining the numerical stability. NF4 was chosen because it allocates its quantization levels according to normal distribution, making it particularly suitable for Transformer models. Moreover, NF4 uses a non-linear mapping of the intervals, such that the densest regions of the distribution preserve a higher precision, reducing the loss of information in the weights and maintaining the performance of the model even at the reduction of the numerical precision, allowing us the storage and processing of the model with limited hardware resources. The original weights $W_0$ remain frozen, so that all the general knowledge and the reasoning capacity remain intact allowing to adapt the model to the specific task of the medical domain, without directly modifying the dense weights and optimizing GPU memory.

We used LoRA fine-tuning technique (Low-Rank Adaptation) (Hu et al., 2022), so that the adaptation of the model's knowledge is possible without directly modifying the dense weights. The LoRA method introduces additional low-rank matrices in certain components of the Transformer architecture, which can be trained. Instead of directly modifying the dense matrices of the pre-trained model, it learns only an approximation for the modification of the weights, significantly reducing the number of parameters that would need to be trained. This approximation is represented by the matrix ΔW which is decomposed into the two trainable low-rank matrices $B$ and $A$:

$$W = W_0 + \Delta W = W_0 + BA \tag{2}$$

where and the low-rank matrices are defined as:

$$B \in R^{d \times r} \tag{3}$$

$$A \in R^{r \times k} \tag{4}$$

where $r$ is the low rank of the matrices, $r \ll \min(d, k)$.

With this approximation, the number of the parameters that must be trained decreases from $d \times k$ to $r \times (d + k)$. The modification of the weights introduced by the LoRA method directly influences the way in which the output of the layer $h$ is computed during the forward pass:

$$h = W_0 x + (\frac{\alpha}{r}) BAx \tag{5}$$

where $x$ represents the input vector, $\alpha$ represents the scaling factor that controls the influence of the LoRA adapter on the initial weights.

At the beginning of the training, the matrix $A$ is initialized with random values extracted from a Gaussian distribution and $B$ with zeros, so that before starting the fine-tuning process the outputs of the model that must be adapted are identical to those of the pre-trained model. During the training, only the parameters of the adapters are updated, while the initial weights remain frozen. We chose r = 32 and $\alpha$ = 32, such that their ratio has the value 1, which allows us a balanced contribution between the general knowledge of the pre-trained model and the specific information from the medical domain that must be learned through the fine-tuning process. We applied a dropout of 0.05 that introduces a regularization mechanism during the training of the LoRA adapters, randomly disabling a part of the connections to prevent overfitting on the medical dataset. The adapters were integrated into both the dense layers and the query, key, and value projection modules of the self-attention mechanism, enabling the model to adapt both feature transformations and token-to-token relationships.

Training was performed using supervised fine-tuning with QLoRA. During this process only matrices $A$ and $B$ were updated and the pretrained model weights $W_0$ remained frozen. The fine-tuning objective is to learn to predict sequentially each token based on the previous context. Therefore, the input text sample is transformed into a sequence of tokens $x = (x_1, x_2, \dots, x_n)$ where each $x_i$ represents a token and the model processes this sequence from left to right and learns to predict each token $x_i$ based on the previous ones $x_j$, where $j < i$.

We use the Cross-Entropy Loss function to measure the performance of the model during training. It penalizes the difference between the generated tokens and the correct ones from the dataset, thus managing to assign higher probabilities to the correct tokens (Radford et al., 2018):

$$L(A, B) = -\sum_{i=1}^{n} \log P\,(x_i \mid x_1, x_2, \dots, x_{i-1}; W_0, A, B) \tag{6}$$

where $x_i$ is the current token is $W_0$ fixed while $A$ and $B$are optimized.

Therefore, the model receives a sequence of tokens, predicts the next token and the difference compared to the correct value is calculated through the loss function, the error being propagated back through backpropagation to update the LoRA parameters, as can be seen in Fig. 3.

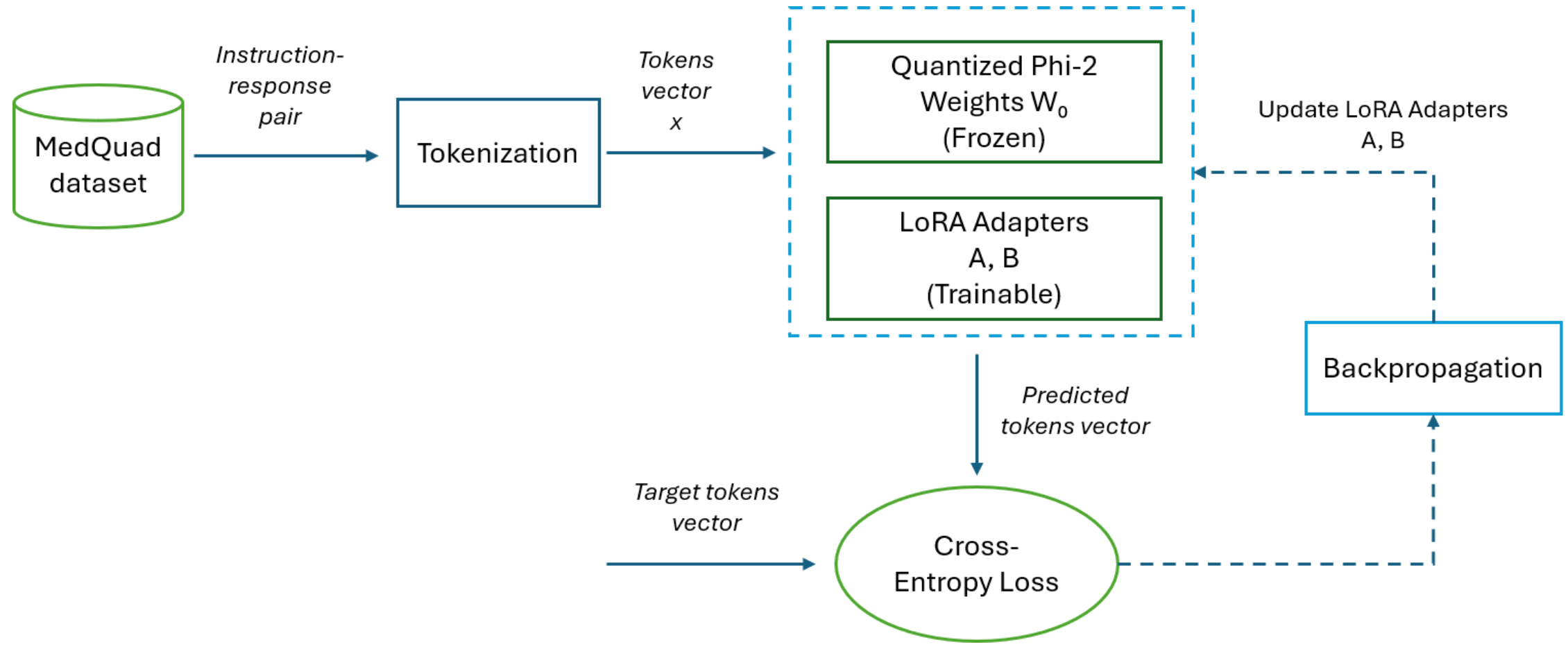

**Figure 3. QLoRA-based supervised fine-tuning process**

To reduce this error progressively, such that we minimize the loss function and update the parameters of the model, it is necessary to use an optimization algorithm. We choose the paged_adamw_8bit optimizer, an efficient variant of the AdamW algorithm (Loshchilov&Hutter, 2017). This represents an improvement of the Adam algorithm by decoupling the regularization term (weight decay) from the gradient updates, thus obtaining a better stability for a training process. One advantage of AdamW is that it manages the problem of GPU memory overflow during sudden variations of the gradient by using a unified memory mechanism, which allows the temporary moving of some states from the GPU to the system RAM when it is necessary.

Additionally, to further reduce the memory consumption, we used several techniques for the optimization of the training process of the model. The batch size was kept at a reduced value and, to compensate for this and to allow a more stable training, we used a gradient accumulation technique with four steps, thus allowing the accumulation of the gradient over multiple iterations before updating the parameters, simulating a larger batch without increasing the GPU consumption. To optimize the moment of the forward pass, the gradient checkpointing technique was enabled, which avoids storing all the intermediate activations.

After the completion of the training, the parameters of the LoRA adapters that incorporate the specific knowledge of the medical domain, obtained using the MedQuad dataset, were saved separately from the base model. In this last stage of the fine-tuning process, we stored the model on disk for later uses, saving both the weights of the LoRA adapters and the tokenizer associated with the model. For inference, the model is reloaded using the base model and the saved LoRA adapters, which we apply dynamically over the original weights.

## *Data augmentation with the specialized SLM*

The SLM model fine-tuned to the medical tasks was used to perform the generative augmentation of the Health Advisor dataset to be used in the training of the LSTM neural network. The original dataset contains vital signs measurements and associated marker labels representing different physiological and pathological states, divided by 9 possible classes: (1) healthy (all values are within normal limits), (2) high blood pressure, (3) low blood pressure, (4) high blood glucose level, (5) low blood glucose level, (6) low oxygen saturation, (7) high body temperature, (8) high heart rate and (9) risk (two or more vital signs are not within normal limits). The distributions from Fig. 4 show that the dataset contains a reduced number of instance values of the vital signs from patients, approximately 600. Also, the instances are imbalanced between the 9 classes. The fewest values are for the class that represents a low value for oxygen saturation (class 6), having 40 entries and the most entries are for the risk class (class 9), having 125 entries.

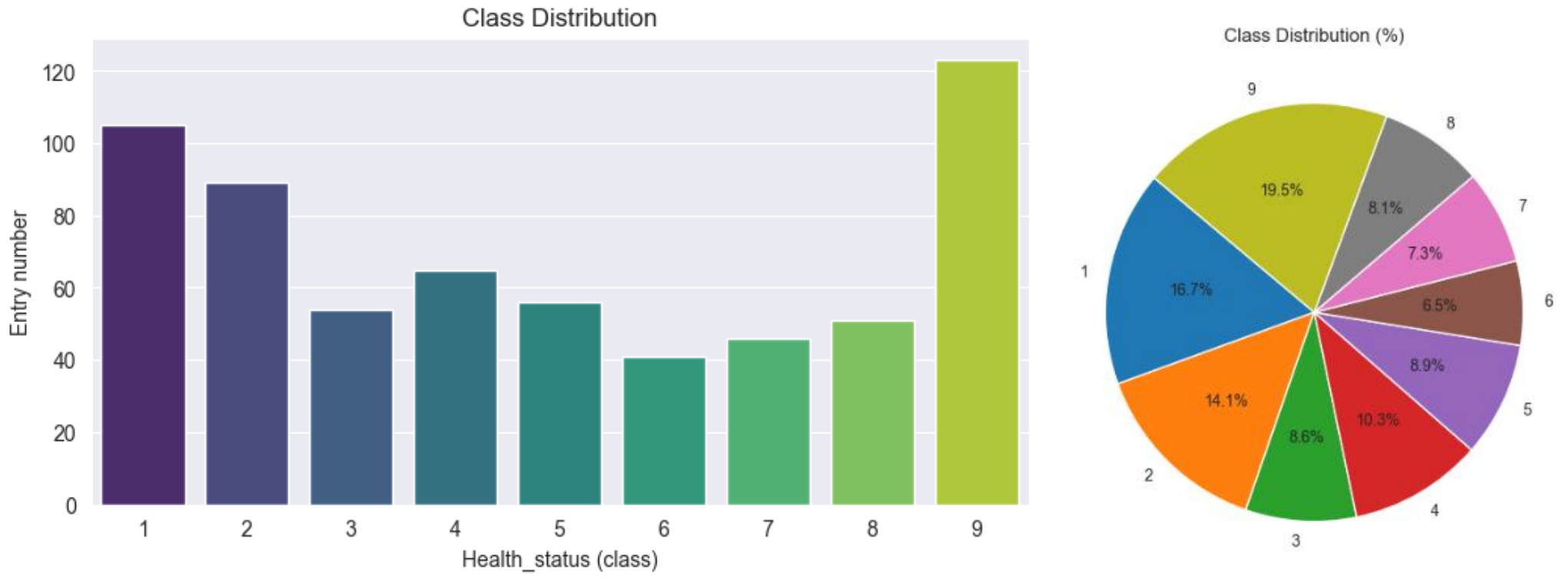

**Figure 4. Class distribution of the initial dataset: (left) absolute values and (right) percentage values**

We use the fine-tuned SLM model to generate new synthetic data instances and address the identified gaps. Fig. 5 presents the augmentation process consisting of the following stages: (i) preprocessing, (ii) prompt construction and the use of few-shot prompting, (iii) data generation using the SLM model, (iv) parsing of the values and (v) postprocessing and validation.

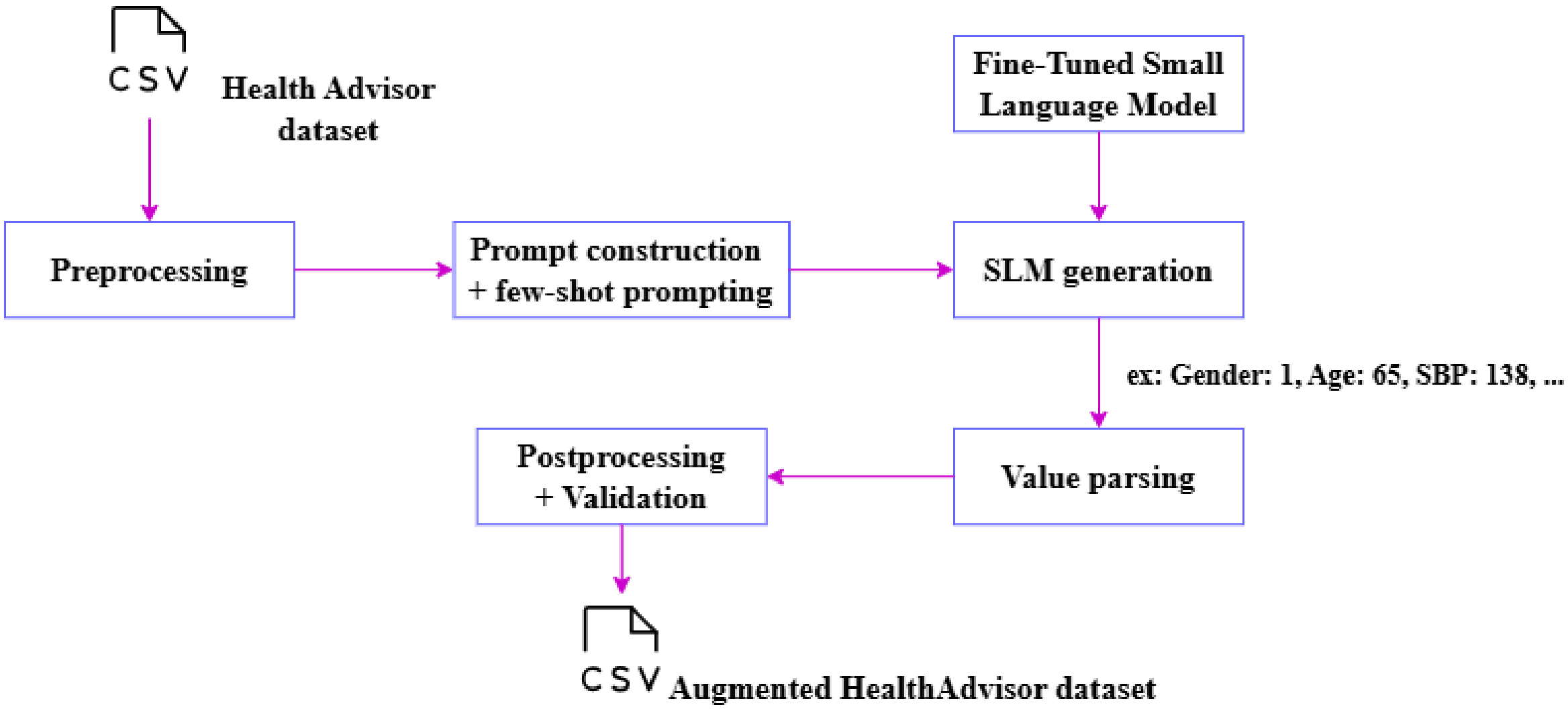

**Figure 5. Data augmentation using the fine-tuning SLM pipeline**

The initial data are mostly in numeric format, with a single exception, the gender of the patient, which was noted with „M” for men and „F” for women. To allow the model to work only with numeric data and to have consistency in the generation of these values, we transformed the uppercase F/M into 0/1. Moreover, to balance the dataset with an equal number of instances, for each of the 9 classes we calculated how many instance values should be added such that we can obtain a uniform distribution. This number for each class was selected by running multiple augmentation scenarios.

A few-shot prompting technique is used with 5 examples for each class that were randomly selected. They are formatted such that it is clear which vital sign it is about. To provide a textual medical information, we also included the class number for which we want to generate data instances together with the physiological state (see Fig. 6).

```
### Instruction: The following are vital signs for patients with Health Status 1 (Healthy).

- Gender (0=F, 1=M): 1, Age: 67, Systolic BP: 138, Diastolic BP: 88, Heart Rate: 98, Glucose: 128, Oxygen Saturation: 97, Temperature: 98.0

- Gender (0=F, 1=M): 0, Age: 59, Systolic BP: 136, Diastolic BP: 85, Heart Rate: 96, Glucose: 125, Oxygen Saturation: 98, Temperature: 97.9

- ...

Generate one new, unique patient entry with similar vital signs following the same format.
```

**Figure 6. prompt example for generating values for class 1 - healthy patient**

Based on the previous prompt, the fine-tuned SLM is employed to generate synthetic patient records containing medically plausible numerical values. Data generation is performed through stochastic decoding mechanisms, including temperature and top-p sampling, which encourage sample diversity while maintaining consistency with the learned data distribution. As a result, the generated instances retain the clinical and statistical properties of the original classes while introducing additional variability. These results are parsed using regular expressions to extract the numerical values for each vital sign, each attribute being identified in the text based on the labels defined in the prompt and, to ensure the compliance with the desired format. We remove the instances that had missing data or were not in numeric format. This parsing allows us to integrate the new instances into the dataset so that they can be used later for neural network training.

Finally, the data was validated by checking the inclusion within physiologically plausible ranges for each vital sign and the newly generated instances that do not meet these conditions were removed. These ranges used were established based on the distribution of the values for each class from the initial dataset to ensure that the obtained data remain consistent with the real ones. Subsequently, the validated instances are transformed into the format required for inclusion in the dataset, the values for the gender of the patient being converted back into the initial representation (from 0/1 into F/M) and then added to the original dataset file. This stage ensured the obtaining of a final dataset that is coherent, valid, with a uniform distribution between classes and compatible with the training pipeline of the LSTM based prediction model.

# Results and discussion

## *Fine-tunned model evaluation*

To evaluate the performance of the fine-tuned model, we performed a quantitative analysis, based on lexical and semantic evaluation metrics by comparing the baseline pre-trained model (Phi-2) to the fine-tuned one. For both models we used the same prompt format, the same inference configuration and the generated responses were compared with the reference responses from the dataset.

The quantitative analysis of the model was performed using an evaluation subset that represents 10% of the MedQuad dataset, the remaining 90% being used previously for the actual fine-tuning process. During inference, each instruction from the evaluation subset was provided to both models using the same instruction-response prompt structure as the one explained previously. The responses provided by them were generated in a deterministic manner, with a maximum length of 350 tokens and a repetition penalty of 1.15 to maintain a consistent comparison between the baseline and fine-tunned SLM.

For the lexical metrics, the actual evaluation was performed by comparing the word sequences (n-grams) between the generated responses. For the semantic evaluation, both the generated responses and the reference ones were transformed into embedding vector representations using the Sentence-BERT all-MiniLM-L6-v2 model (Reimers & Gurevych, 2019), the similarity between them being calculated based on the proximity between these vectors, using cosine similarity. We have used a set of metrics, both lexical and semantic, allowing the evaluation of the structural correctness of the responses and their relevance from a medical point of view. For the lexical similarity we used the ROUGE (Lin, 2004) metrics, specifically Rouge-1, Rouge-2, Rouge-L and Rouge-Lsum. This metric is widely used in the literature to measure the degree of coverage of information from the reference text, highlighting how much of the relevant content is captured in the generated responses. Rouge-L and Rouge-Lsum allow the evaluation of the coherence and the global structure of the entire text. Regarding the evaluation of semantic similarity, we used METEOR (Banerjee & Lavie, 2005), cosine similarity and BERTScore (Zhang et al., 2019) metrics. METEOR extends the lexical evaluation by considering morphological variations and the order of words, providing an evaluation closer to human interpretation compared to ROUGE. Cosine similarity measures the global proximity between the vector representations of the texts, while BERTScore provides a deeper semantic evaluation, comparing the tokens and capturing the relationships between them. The use of both lexical and semantic metrics allows a complete evaluation of the model from the point of view of linguistic structure, as well as the understanding of the generated responses. Figure 7 shows a comparative view for the metrics results of the base and fine-tuned models.

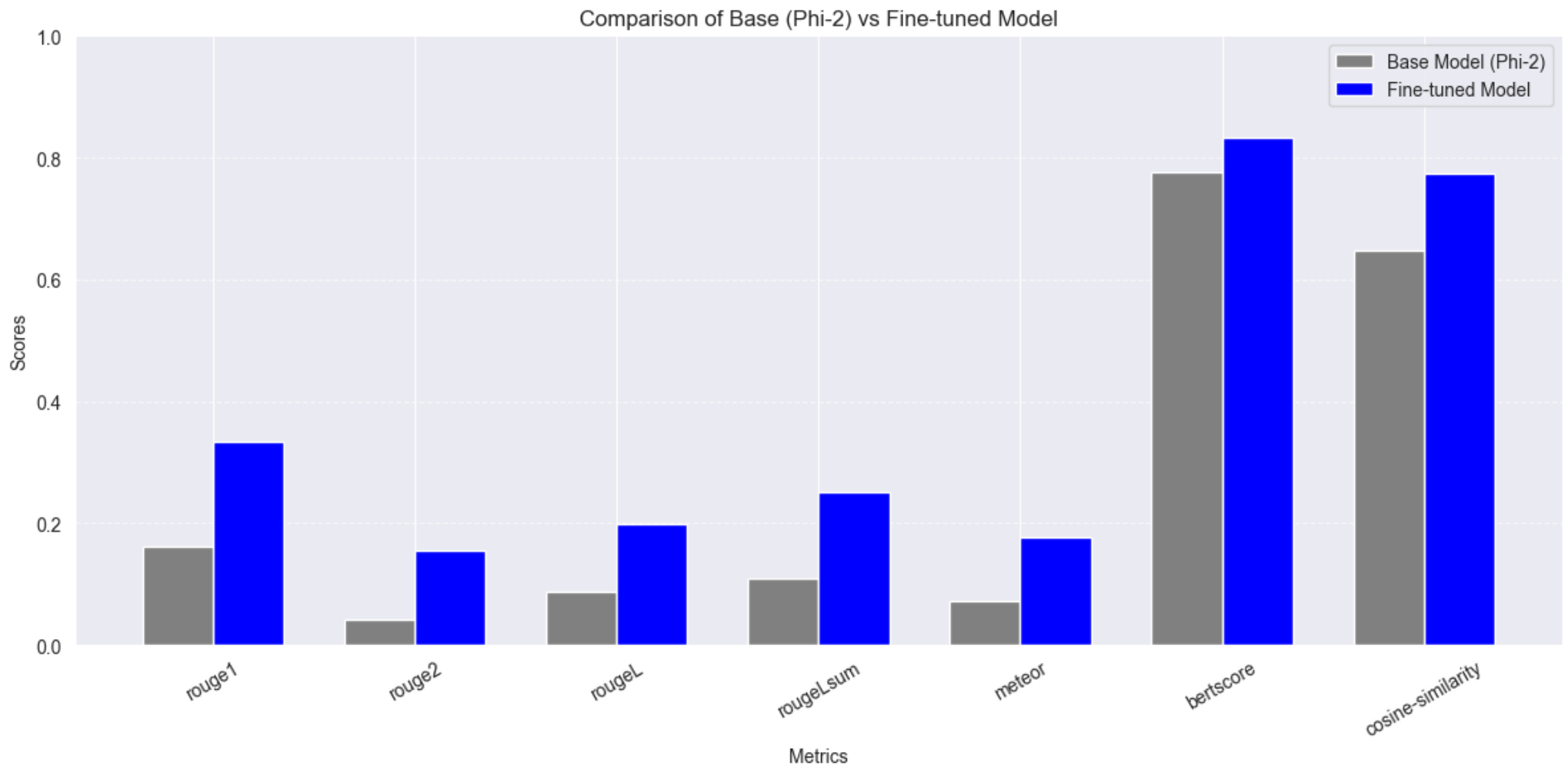

**Figure 7. Comparison between the pretrained model and the fine-tuned model**

Detailed results obtained for this evaluation are shown in Table 1 and they highlight a significant improvement in the performance of the fine-tuned model compared to the pretrained model across all the used metrics. Regarding lexical similarity, the ROUGE scores indicate a better matching of the structure and the content of the generated responses with the reference ones, managing to produce word sequences that are more coherent and better contextually aligned.

**Table 1. Comparison between the pretrained model and the fine-tuned model**

| *Metric* | *Base model* | *Fine-tuned model* | *Improvement (%)* |
|---|---|---|---|
| *ROUGE-1* | 0.162 | 0.333 | 105.35% |
| *ROUGE-2* | 0.042 | 0.155 | 268.86% |
| *ROUGE-L* | 0.087 | 0.198 | 128.03% |
| *ROUGE-LSum* | 0.109 | 0.250 | 129.09% |
| *METEOR* | 0.073 | 0.176 | 140.65% |
| *Cosine Similarity* | 0.647 | 0.773 | 19.50% |
| *BERTScore* | 0.777 | 0.833 | 7.19% |

The improvements observed for the METEOR, cosine similarity and BERTScore metrics confirm the fact that the model not only reproduces better the structure of the responses, but also understands better the meaning of the instructions, being able to generate more semantically relevant responses. The largest increases were for ROUGE-2, this indicating an efficient adaptation to the instruction-response format specific to MedQuad dataset, however, the increase of the semantic metrics scores also confirms the fact that the model has learned the relationships between the medical concepts, generating more contextually correct responses.

In addition to the evaluation based on the previous metrics, we performed a qualitative manual analysis of the responses generated by the pretrained model and by the one obtained following the fine-tuning process based on a subset of examples randomly selected from the evaluation set. Comparing the generated responses, it was observed that the pretrained base model tends to provide more general responses, sometimes incomplete or less relevant for the specific medical context of the instruction, while the fine-tuned model produces responses that are better structured and aligned with the medical content of the instructions, reflecting the knowledge acquired from the MedQuad dataset. This evaluation confirms the results obtained through the automatic metrics, highlighting the fact that the fine-tuning process contributed to the adaptation of the model to the tasks from the medical domain.

## ***Data augmentation method evaluation***

To assess the effectiveness of the data augmentation approach, we retrained the LSTM model previously developed for vital sign prediction using the newly augmented HealthAdvisor dataset. The obtained results were subsequently compared with the baseline performance reported in our earlier work (Burlacu & Anghel, 2024). The model is a recurrent neural network formed of 3 LSTM layers, 4 dropout layers, 3 batch normalization layers and 2 dense layers. The output offers a multiclass classification into 9 possible categories, framing the analysed values into different physiological and pathological states (e.g.: high / low blood pressure, high temperature, low oxygen saturation, etc.).

To establish the baseline the LSTM model was evaluated on the original HealthAdvisor dataset using accuracy, precision, recall and F1-score as metrics. As illustrated in Figure 8, we obtained an accuracy of 0.89 and a value for the cross-entropy loss function of 0.84.

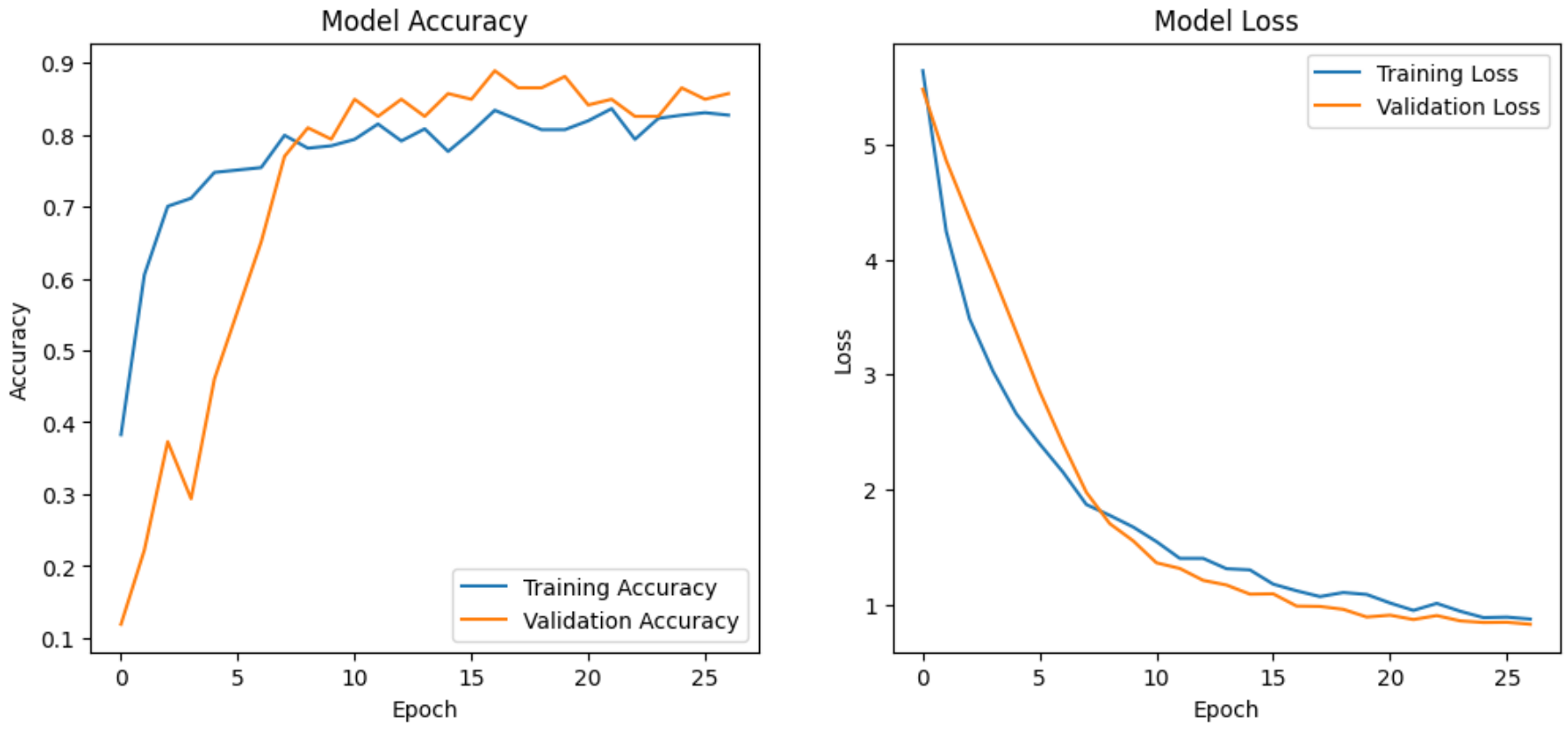

**Figure 8. Accuracy and the loss function of the LSTM model before augmentation**

In Figure 9 are illustrated the values obtained for precision, recall and F1-score. Regarding precision, the average of the values obtained for each class gave us a score of 0.88, for recall the average was 0.89 and for F1-score the average value was 0.88, the best score is 1.0 for the class that signifies increased body temperature and the lowest value is approximately 0.8 for the last class.

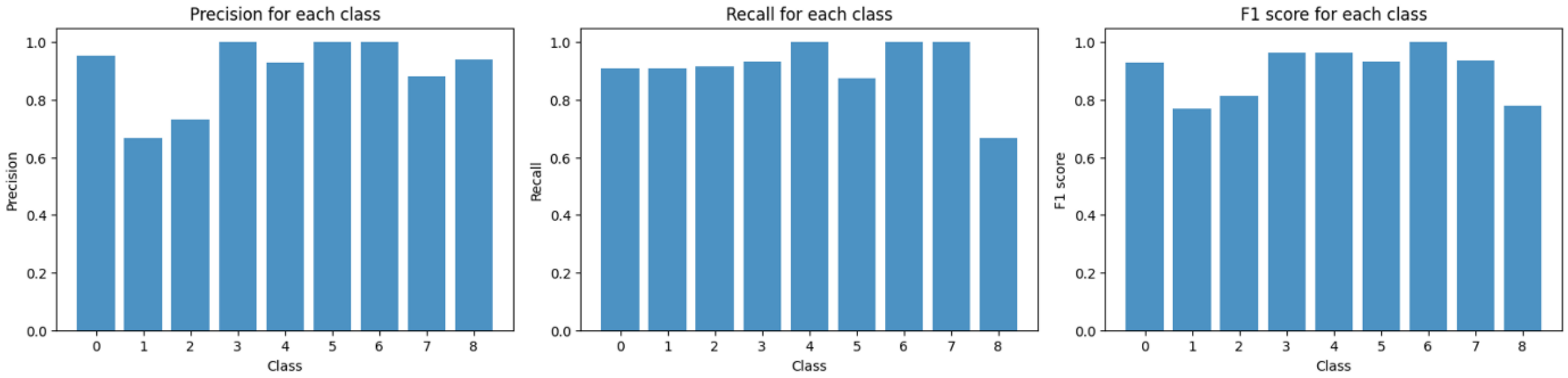

**Figure 9. Precision, recall, F1-score of the LSTM model before augmentation**

To determine the improvements brought by the augmentation, we define three scenarios in which the LSTM model was retrained aiming to also evaluate the relationship between the volume of the augmented data and the performance of the prediction model.

In *the first scenario*, the HealthAdvisor dataset was augmented so that each class contained at least 120 instances. This threshold was chosen to achieve a nearly balanced class distribution, while limiting the amount of synthetic data generated, given that the largest class in the original dataset contained 125 instances. In this way we evaluate the impact of a low augmentation as we added only approximately 400 entries over the initial dataset. As results of better data balancing, an improvement in performance was observed, as is also illustrated in Figure 10. The accuracy increased to 91%, representing an improvement of approximately 2 percentage points over the initial result. The loss decreased steadily, reaching 0.69 compared to the initial value of 0.84. The average precision across all classes improved to 0.92, an increase of approximately 4 percentage points, with most classes now achieving scores above 0.90. Similarly, the recall values for most classes approached 1.0, except for the last class, which remained slightly above 0.60. The average F1-score also increased to 0.92, corresponding to an improvement of 4 percentage points. This enhancement was particularly noticeable for the second and third classes, resulting in a more uniform performance across the 9 categories.

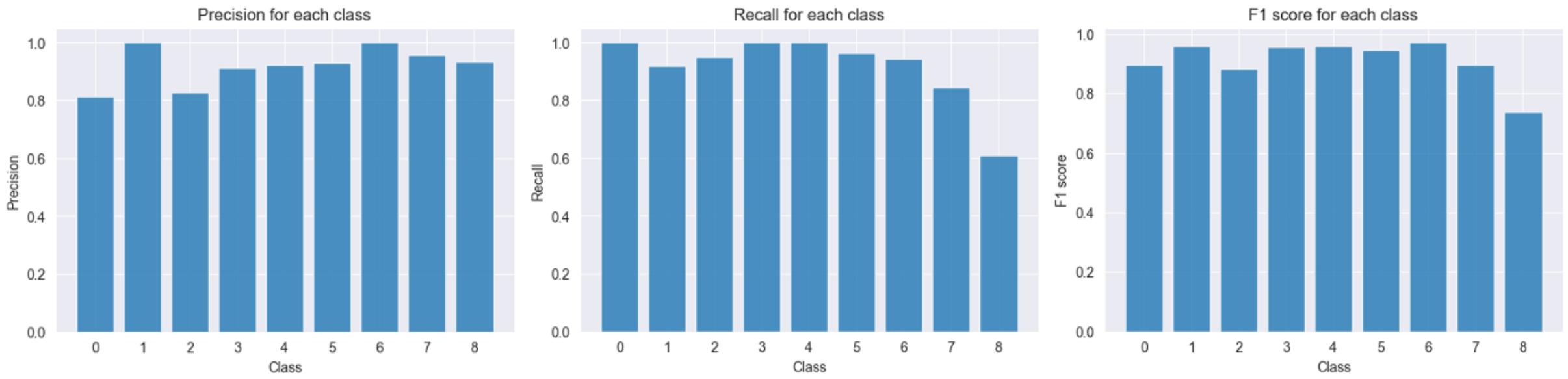

**Figure 10. Precision, recall, F1-score for the low augmentation scenario**

The *second scenario* involved moderate augmentation of the dataset so that each class contained 250 instances. Although this led to a further improvement in performance, the gains were smaller than those observed in the first scenario, as shown in Figure 11. The accuracy increased to 92%, while the loss decreased further to 0.59. The precision, recall, and F1-score also showed modest improvements. Notably, the last class, which previously exhibited relatively low recall and F1-score values, achieved scores close to 0.8 for both metrics.

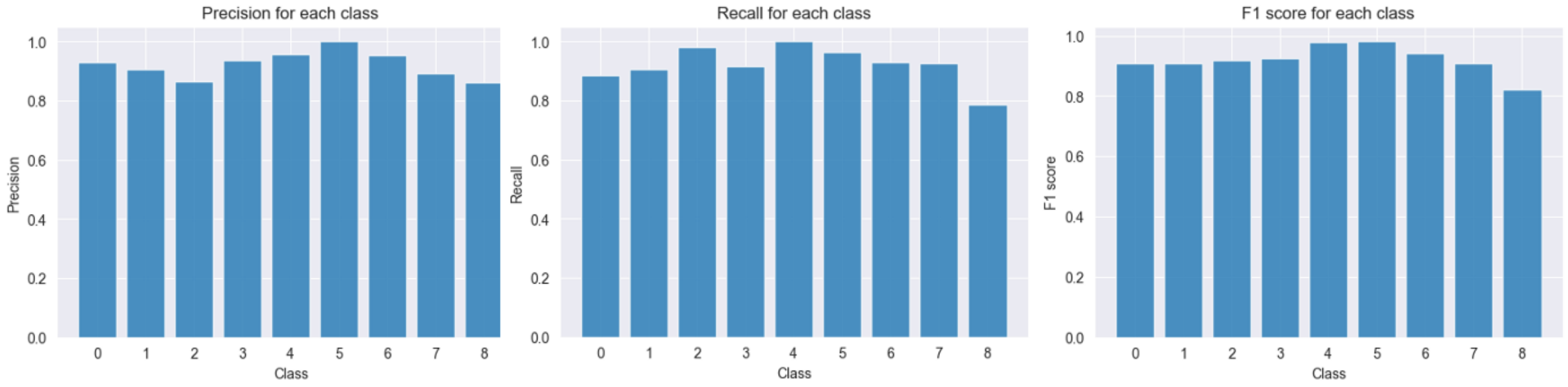

**Figure 11. Precision, recall, F1-score for the moderate augmentation scenario**

In the *third scenario*, the dataset was heavily augmented so that each of the 9 classes contained 500 instances. After retraining the LSTM model on the augmented dataset, a further improvement in performance was observed, more pronounced than in the previous scenario, as illustrated in Figure 12. The accuracy increased to approximately 94%, while the loss decreased to 0.49. The precision, recall, and F1-score values also improved for most classes, including those that had previously exhibited lower performance. As a result, the differences between classes were reduced, indicating a more uniform distribution of predictive performance across all categories.

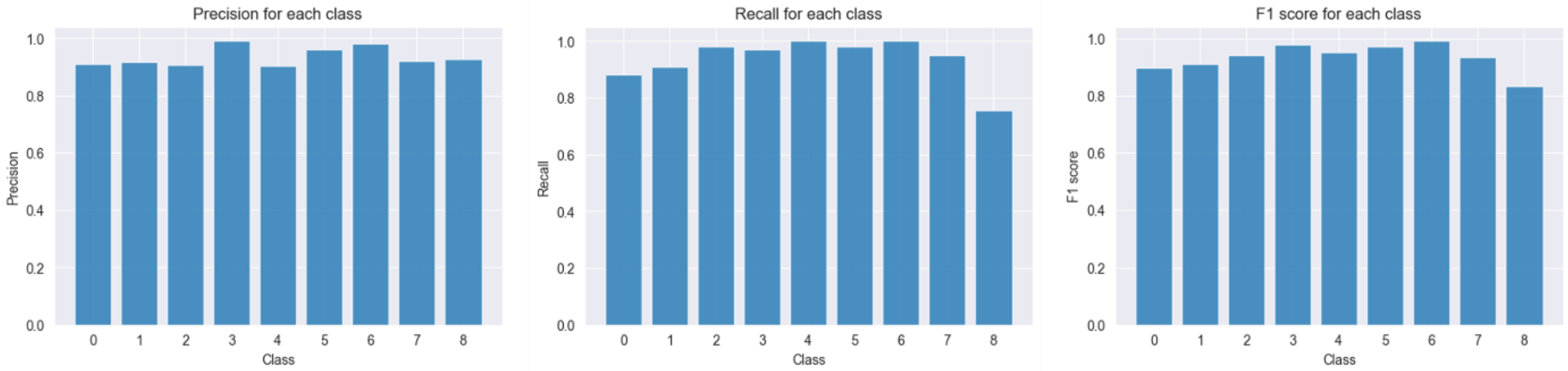

**Figure 12. Precision, recall, F1-score for the heavy augmentation scenario**

The results obtained across the three analyzed scenarios clearly demonstrate the positive impact of data augmentation on the performance of the LSTM model. Compared to the model trained on the original, unmodified dataset, all augmented datasets led to consistent improvements across the evaluated performance metrics. (see Table 2).

**Table 2. Comparison of the values obtained depending on the executed scenario**

| *Scenario* | *Accuracy* | *Loss* | *Precision* | *Recall* | *F1-score* |
|---|---|---|---|---|---|
| *Baseline* | 0.8889 | 0.8429 | 0.88 | 0.89 | 0.88 |
| *Augmented 120 / class* | 0.9171 | 0.6914 | 0.92 | 0.93 | 0.92 |
| *Augmented 250 / class* | 0.9200 | 0.5981 | 0.93 | 0.94 | 0.93 |
| *Augmented 500 / class* | **0.9367** | **0.4989** | **0.94** | **0.95** | **0.94** |

# Conclusions

In this paper, a method of clinical data augmentation was proposed using a SLM, Microsoft Phi-2 adapted to the medical domain through fine-tuning based on the MedQuad dataset and using QLoRA, with the purpose of being able to better understand medical terminology and context. The model was used to generate structured synthetic instances for the HealthAdvisor dataset, which contains values of vital signs and classes associated with pathological or physiological states, these generated data being later used for retraining an LSTM model used in the analysis of vital signs.

The experimental evaluation demonstrated the effectiveness of the proposed approach. The fine-tuned SLM outperformed the pretrained model according to both lexical metrics, such as ROUGE, and semantic metrics, such as BERTScore, confirming the benefits of domain adaptation for medical text generation. The results obtained with the LSTM prediction model showed that synthetic data augmentation consistently improved predictive performance across all analyzed scenarios. Increases were observed for accuracy, precision, recall, and F1-score, with the best results achieved when the dataset was balanced to 500 instances per class. In this configuration, the accuracy reached 93.7%, compared to the baseline value of 89%. These results demonstrate that an SLM adapted to the medical domain through fine-tuning can serve not only as a tool for question-answering tasks but also as an effective mechanism for generating high-quality synthetic medical data. Nevertheless, the proposed approach presents several limitations. Although the generated data underwent a validation process, the SLM may occasionally produce inconsistent or less realistic instances, particularly when extrapolating beyond the patterns observed during training. Furthermore, the substantial increase in dataset size results in higher computational costs during model training. Future work may therefore focus on enhancing the validation of synthetic data and investigating more computationally efficient augmentation strategies.

### Ethics statement and consent to participate

Given that the research uses open access datasets, no ethical approval was necessary.

### Consent for publication

Not applicable.

### Availability of data and material

The datasets generated during and/or analysed during the current study are available from the corresponding author on reasonable request.

### Competing interest

The authors declare no competing financial or non-financial interests.

**Funding**

This work was supported by a grant from the Ministry of Research, Innovation and Digitization, CNCS/CCCDI - UEFISCDI, project number ERANET-PARTENERIAT-THCS-TransCare-1, within PNCDI IV.

**Authors' contributions**

Conceptualization, M.-S. B., I.A. and T.C.; Methodology, M.-S. B., I.A. and T.C.; writing—original draft preparation, M.-S. B., I.A. and T.C.; writing—review and editing, I.A. and T.C.; All authors read and agreed to the submitted version of the manuscript.

**Acknowledgements**

Not applicable.